\pdfoutput=1

\documentclass[11pt]{article}

\usepackage[]{acl}

\usepackage{times}
\usepackage{latexsym}
\usepackage{subfigure}

\usepackage[greek,english]{babel}
\usepackage{alphabeta}
\usepackage{graphics}
\usepackage{graphicx}
\usepackage{rotating}
\usepackage{multirow}
\usepackage{adjustbox}
\newcommand*\rot{\rotatebox{90}}
\usepackage{array,booktabs}
\usepackage{float}
\usepackage{amsmath}
\usepackage{amssymb}
\usepackage{pifont}%
\newcommand{\cmark}{\ding{51}}%
\newcommand{\xmark}{\color{lightgray}\ding{55}}%
\usepackage{bm}

\usepackage{comment}
\usepackage{microtype}

\usepackage{fixltx2e}

\usepackage{array}
\usepackage{colortbl}

\usepackage{url}

\usepackage{pgfplots}
\usepackage{contour}
\usepackage{xspace}
\usepackage{paralist}
\pgfplotsset{width=10cm,compat=1.9} 

\usepackage{algorithm,algpseudocode}
\usepackage{amsmath}
\usepackage{amsfonts}
\usepackage{amssymb}

\definecolor{darkblue}{rgb}{0,0,.5}
\definecolor{darkgreen}{rgb}{0,.5,0}
\definecolor{lightgray}{rgb}{.8,.8,.8}
\definecolor{aliceblue}{rgb}{0.75, 0.75, 1.0}
\definecolor{darkseagreen}{rgb}{0.46, 0.74, 0.46}
\definecolor{alizarin}{rgb}{0.82, 0.1, 0.26}
\definecolor{airforceblue}{rgb}{0.36, 0.54, 0.66}
\definecolor{red_graph}{rgb}{0.98, 0.8, 0.8}
\definecolor{blue_graph}{rgb}{0.8, 0.98, 0.8}
\definecolor{red}{rgb}{0.8, 0.0, 0.0}
\definecolor{burgundy}{rgb}{0.5, 0.0, 0.13}
\definecolor{britishracinggreen}{rgb}{0.0, 0.26, 0.15}
\usepackage[inline]{enumitem}

\newcommand{\bleu}{\textsc{bleu}\xspace}

\newcommand{\laser}{\textsc{laser}\xspace}

\newcommand{\ted}{\textsc{ted}\xspace}
\newcommand{\nmt}{\textsc{nmt}\xspace}

\newcommand{\en}{\textsc{en}\xspace}
\newcommand{\el}{\textsc{el}\xspace}
\let\emptyset\varnothing

\newcommand{\dataviz}[3]{\begin{tikzpicture}[scale=1.5]
\filldraw[draw=darkgreen,fill=darkgreen!30] (#2,0) rectangle (#1+#2+#3,0.15); 
\filldraw[draw=alizarin, fill=alizarin!30] (#1,0) rectangle (#1+#2,0.15);
\filldraw[draw=airforceblue, fill=airforceblue!30] (0,0) rectangle (#1,0.15);
\end{tikzpicture}}

\contourlength{0.2pt}
\newcommand{\hz}{\vphantom{\parbox[c]{0.08cm}{\rule{0.08cm}{0.19cm}}}}
\newcommand{\da}{%
  \colorbox{burgundy!30}{\hz\tiny{$\downarrow$}}%
}
\newcommand{\ua}{%
  \colorbox{britishracinggreen!30}{\hz\tiny{$\uparrow$}}%
}
\newcommand{\ph}{%
  \colorbox{white}{\hz\color{white}\tiny{$\downarrow$}}%
}
\newcommand{\phh}{%
  \colorbox{white}{\color{white}$^*$}%
}

\usepackage{color, colortbl}
\newcolumntype{o}{>{\columncolor{gray!10}}c}
\newcolumntype{m}{>{\columncolor{gray!30}}c}
\newcolumntype{h}{>{\columncolor{gray!50}}c}

\newcommand{\dto}{%
  \colorbox{airforceblue!30}{\hz $\mathcal{O}$}%
}
\newcommand{\dtf}{%
  \colorbox{alizarin!30}{\hz $\mathcal{F}$}%
}
\newcommand{\dtb}{%
  \colorbox{darkgreen!30}{\hz $\mathcal{B}$}%
}
\usepackage{xcolor}

\newcommand{\mycomment}[3]{}

\newcommand{\ignore}[1]{}
\usepackage[T1]{fontenc}

\usepackage[utf8]{inputenc}

\usepackage{microtype}

\title{Can Synthetic Translations Improve Bitext Quality?}

\author{Eleftheria Briakou \normalfont{and}  \textbf{Marine Carpuat} \\
  Department of Computer Science \\
  University of Maryland\\
  College Park, MD $20742$, USA\\
  \texttt{\href{mailto:ebriakou@cs.umd.edu}{ebriakou@cs.umd.edu}, \href{mailto:marine@cs.umd.edu}{marine@cs.umd.edu}}} 

\begin{document}
\maketitle
\begin{abstract}

Synthetic translations have been used for a wide range of \textsc{nlp} tasks primarily as a means of data augmentation. This work explores, instead, how synthetic translations can be used to \textit{revise} potentially imperfect reference translations in mined bitext. We find that synthetic samples can improve bitext quality without any additional bilingual supervision when they replace the originals based on a semantic equivalence classifier that helps mitigate \textsc{nmt} noise. The improved quality of the revised bitext is confirmed intrinsically via human evaluation and extrinsically through bilingual induction and \textsc{mt} tasks.

\end{abstract}

\section{Introduction}

While human-written data remains the gold standard to train Neural Machine Translation (\nmt) and Multilingual \textsc{nlp} models, there is growing evidence that synthetic bitext samples---sentence-pairs that are translated by \nmt---benefit a wide range of tasks. They have been used to enable semi-supervised  \textsc{mt} training from monolingual data ~\cite{sennrich-etal-2016-improving, zhang-zong-2016-exploiting, hoang-etal-2018-iterative}, to induce bilingual lexicons~\cite{artetxe-etal-2019-bilingual, shi-etal-2021-bilingual}, and to port models trained on one language to another~\cite{conneau-etal-2018-xnli, yang-etal-2019-paws}. 

While synthetic bitexts are useful additions to original training data for downstream tasks, it remains unclear how they differ from naturally occurring data. Some studies suggest that synthetic samples might be simpler and easier to learn~\citep{chuntin, xu-etal-2021-distilled}. Recognizing that naturally occurring bitext can be noisy, 
for instance, when they are mined from comparable monolingual corpora~\cite{resnik-smith-2003-web,fung-yee-1998-ir,espla-etal-2019-paracrawl, schwenk-etal-2021-wikimatrix}, we hypothesize that synthetic bitext might also directly improve the equivalence of the two bitext sides. Thus synthetic samples might be useful not only for data augmentation but also to revise potentially noisy original bitext samples.

In this paper, we present a controlled empirical study comparing the quality of bitext mined from monolingual resources with a synthetic version generated via \textsc{mt}. 
We focus on the widely used WikiMatrix bitexts for a distant (i.e, \textsc{en-el}) and a similar language-pair (i.e, \textsc{en-ro}), since it has been shown that this corpus contains a significant proportion of erroneous translations \cite{audit_multi}. 
We generate synthetic bitext by translating the original training samples using \textsc{mt} systems trained on the bitext itself and therefore do not inject any additional supervision in the process. We also consider selectively replacing original samples with forward and backward synthetic translations based on a semantic equivalence classifier, which is also trained without additional supervision.

We show that the resulting synthetic bitext improves the quality of the original intrinsically using human assessments of equivalence and extrinsically on bilingual induction (\textsc{bli}) and \textsc{mt} tasks. We present an extensive analysis of synthetic data properties and of the impact of each step in its generation process. This study brings new insights into the use of synthetic samples in \textsc{nlp}. First, intrinsic evaluation shows that synthetic translations, in addition to ``normalizing'' the bitext~\citep{chuntin, xu-etal-2021-distilled}, 
could potentially provide reference translations that are more semantically equivalent to the source than the original ones.

Furthermore, the improved bitext provides more useful signals for \textsc{bli} tasks and \nmt training in two settings (training from scratch; continued training), as confirmed by our extrinsic evaluations. Finally, ablation analyses that compare different ways to combine synthetic translations show that using \textit{both translation directions} and \textit{filtering using semantic equivalence} is key to improving bitext quality and calls for further exploration of best practices for using synthetic translations in \textsc{nlp} tasks.

\section{Background}
\paragraph{Synthetic Translations} Generating synthetic translations
has mainly been studied as a means of data augmentation for \textsc{nmt} through forward translation~\cite{zhang-zong-2016-exploiting} or back-translation~\cite{sennrich-etal-2016-improving, marie-etal-2020-tagged} of monolingual resources.
Moreover, recent lines of work use synthetic translations to augment the original parallel data: \citet{NEURIPS2020_7221e5c8} diversify the parallel data via translating both sides using multiple models and then merging them with the original to train a final \textsc{nmt} model;
\citet{jiao-etal-2020-data} employ a similar approach to rejuvenate inactive examples that contribute the least to the model performance. Sequence-level knowledge distillation \citep{kim-rush-2016-sequence}  can also be viewed as replacing original bitext with synthetic translations.  While its original goal was to guide the training of a student model of small capacity with the output of a teacher of high capacity, distillation is also necessary to effectively train some categories of \textsc{mt} architectures such as non-autoregressive models \citep{jiatao}. While it is not entirely clear why synthetic distilled samples are superior to original bitext in this case, recent studies suggest that the synthetic samples are simpler and thus easier to learn from \citep{chuntin, xu-etal-2021-distilled}. 
\paragraph{Synthetic Data Selection} Prior work covers 
a wide spectrum of different selection strategies on top of synthetic translations generated from monolingual samples. Each of them focuses on identifying samples with specific properties:
\citet{axelrod-etal-2011-domain} sample sentences that are most relevant to a target domain with the goal of creating pseudo in-domain bitext; 
\citet{hoang-etal-2018-iterative} generate synthetic parallel data iteratively from increasingly better back-translation models for improving unsupervised \textsc{nmt};
\citet{fadaee-monz-2018-back} focus on the diversity of synthetic samples and 
sample synthetic translations containing words that are difficult to predict using prediction losses and frequencies of words.
By contrast, our empirical study investigates whether synthetic translations can be used to \textit{\textbf{selectively replace}} original references to improve bitext quality rather than augmenting it.
\paragraph{Bitext Quality}  
Mining bitext from the web results in large-scale corpora that are usually collected without guarantees about their quality. For instance, they contain noisy samples, ranging from untranslated sentences to sentences with no linguistic content~\cite{khayrallah-koehn-2018-impact,caswell-etal-2020-language}. Some of this noise is typically filtered out automatically using heuristics~\cite{ramirez-sanchez-etal-2020-bifixer} or \textsc{nmt} model scores~\cite{junczys-dowmunt-2018-dual, koehn-etal-2019-findings}. Yet, even after this noise filtering, a wide range of the remaining samples contains fine-grained semantic divergences~\cite{briakou-carpuat-2020-detecting}. Our past work explored strategies to mitigate the impact of these divergences on \textsc{mt} models by incorporating divergence tags as token-level factors \cite{briakou-carpuat-2021-beyond}, and designing an approach to automatically edit divergent samples with noisy supervision from monolingual resources \cite{bitextedit}.
By contrast, this work explores whether synthetic translations can be used to replace potentially fine-grained divergences using only the bitext we seek to revise.

\section{Approach}

\begin{algorithm}[!t]
\small
\caption{Revising Bitext: Given a bitext $\mathcal{D}=(S,T)$, a divergent scorer $\mathbf{R}$, and a margin score $t$, return revised bitext $\tilde{\mathcal{D}}$}\label{alg:algo}

\begin{minipage}{0.4\textwidth}
\begin{algorithmic}[1]
        \Procedure{\textsc{train($\mathcal{D}=(S,T)$)}}{}
        \State Train $M_{S \rightarrow T}$ on $\mathcal{D}$ until convergence
        \State \textbf{return} $M_{S \rightarrow T}$
        \EndProcedure
\end{algorithmic}
\end{minipage}
\hfill

\begin{algorithmic}[1]


    \Procedure{Equivalize}{$\mathcal{D}=(S,T)$} 
    \State $M_{S \rightarrow T}$ $\leftarrow$
    \textsc{train($\mathcal{D}=(S,T)$)}
    
    \State $M_{T \rightarrow S}$ $\leftarrow$
    \textsc{train($\mathcal{D}=(T,S)$)}
    
    \State $\tilde{\mathcal{D}} \leftarrow \emptyset$
        \For{$i$ $\in$ $1$,...,$\vert \mathcal{D} \vert$}
        
        \State ($S_i$, \colorbox{green!30}{\hz {$\hat{T_i}$}}) $\leftarrow$ ($S_i$, $M_{S\rightarrow T}(S_i)$)
        \State (\colorbox{yellow!40}{\hz$\hat{S_i}$}, $T_i$) $\leftarrow$ ($M_{T\rightarrow S}(T_i), T_i$)

        \State $d_{F} \leftarrow$ $\mathbf{R}$\textbf{(}$S_i$, \colorbox{green!30}{$\hat{T_i}$}\textbf{)}
        $-$ $\mathbf{R}$\textbf{(}$S_i$ ,$T_i$\textbf{)} 
        \State $d_{B} \leftarrow$ $\mathbf{R}$\textbf{(}\colorbox{yellow!40}{$\hat{S_i}$},$ T_i$\textbf{)}
        $-$ $\mathbf{R}$\textbf{(}$S_i$,$T_i$\textbf{)} 
        
        \If{$\max$($d_{F}$, $d_{B}$) $>$ $t$}
            \If{$\max=d_{F}$}
                \State $\tilde{\mathcal{D}}$ $\leftarrow$ $\tilde{\mathcal{D}}$ $\cup$ \{($S_i$, {\colorbox{green!30}{$\hat{T_i}$}})\}
            \Else
                \State $\tilde{\mathcal{D}}$ $\leftarrow$ $\tilde{\mathcal{D}}$ $\cup$ \{(\colorbox{yellow!40}{$\hat{S_i}$}, $T_i$)\}
            \EndIf
        \Else
            \State $\tilde{\mathcal{D}}$ $\leftarrow$ $\tilde{\mathcal{D}}$ $\cup$ \{($S_i$, $T_i$)\}
        \EndIf

        \EndFor
    \State \textbf{return} $\tilde{\mathcal{D}}$
    \EndProcedure
\end{algorithmic}\label{algo:algo}
\end{algorithm}

This section describes the methods and data we use to produce revised bitexts for our empirical study. 

\subsection{Methods for Revising Bitext}\label{sec:equivalize_NMT} 
We rely on established techniques that can be applied using only the bitext that we seek to revise. First, we train \nmt models on the original bitext to translate in both directions. For each original sentence-pair, we generate a pool of synthetic translations using \nmt and apply a divergence ranking criterion to decide whether and how to replace the original references with a better translation. Algorithm~\ref{alg:algo} gives an overview of the process, and we describe each step below.
\paragraph{Generating synthetic translations} We train \nmt models $M_{S \rightarrow T}$ and $M_{T \rightarrow S}$ on the original bitext to translate in each direction~(lines~$2$-$3$). For each sentence-pair,  they are used to generate two candidates for replacement
by forward and backward translation~(lines~$6$-$7$): ($S_i,M_{S \rightarrow T}(S_i)$) and ($M_{T \rightarrow S}(T_i),T_i$). As a result, \nmt models translate the exact same data that they are trained on. We thus expect translation quality to be high
, and that local errors in the original bitext might be corrected by the translation patterns learned by \nmt models on the entire corpus.

\paragraph{Selective Replacement} We propose to replace an original pair by a candidate \textit{only if}
 the candidate is predicted to better convey the meaning of the source than the original, which we refer to as the \textit{semantic equivalence condition}. We implement this by 
 ranking the original sample ($S_i,T_i$), its revision by forward translation ($S_i,M_{S \rightarrow T}(S_i)$) and its revision by back-translation ($M_{T \rightarrow S}(T_i),T_i$), according to their degree of semantic equivalence. If none of the synthetic samples score higher than the original, it is not replaced~(line~$17$). Otherwise, the original is replaced by the highest scoring synthetic sample  (lines~$10$-$15$). As a result the cardinality of the bitext remains constant. 
 The semantic equivalence condition ($d_{F}$ and $d_{B}$ (lines~$8$-$9$)) is implemented using divergentm\textsc{bert}, a divergent scorer introduced in our prior work \cite{briakou-carpuat-2020-detecting} that is trained on synthetic samples generated by perturbations of the original bitext (e.g., deletions, lexical or phrasal replacements) performed without any bilingual information.

\subsection{Experimental Set-Up}\label{sec:setup_human}

\paragraph{Bitext} We evaluate the use of synthetic translations for revising bitext on two language pairs of the WikiMatrix corpus~\cite{schwenk-etal-2021-wikimatrix}. WikiMatrix consists of sentence-pairs mined from Wikipedia pages using language agnostic sentence embeddings (\laser)~\cite{laser}. Prior work indicates that, as expected, the corpus as a whole comprises many samples that are not exact translations: \citet{audit_multi} report that for more than half of the audited low-resource language-pairs, mined pairs are on average misaligned; \citet{briakou-carpuat-2020-detecting} find that $40\%$ of a random sample of the English-French bitext are not semantically equivalent, and include fine-grained meaning differences in addition to alignment noise.  
We focus on bitexts with fewer than one million sentence pairs in Greek$\leftrightarrow$English (\textsc{el}$\leftrightarrow$\textsc{en}, with $750{,}585$ pairs) and 
Romanian$\leftrightarrow$English (\textsc{ro}$\leftrightarrow$\textsc{en}, with $582{,}134$ pairs), because improving bitext is particularly needed in this data regime. In much higher resource settings, filtering strategies might be sufficient as there might be more high quality samples overall. In much lower resource settings, the data is likely too noisy or too small to effectively revise bitexts using \nmt.
We filter out noisy pairs in the training data using 
\texttt{bicleaner}~\cite{ramirez-sanchez-etal-2020-bifixer} so that our empirical study excludes the most obvious forms of noise, and focuses on the harder case of revising samples that standard preprocessing pipelines consider to be clean.\footnote{\url{https://github.com/bitextor/bicleaner}}

\paragraph{Preprocessing} We use Moses~\cite{koehn-etal-2007-moses} for punctuation normalization, true-casing, and tokenization. We learn $32$K \textsc{bpe}s~\cite{sennrich-etal-2016-neural} per language using \texttt{subword-nmt}~\footnote{\url{https://github.com/rsennrich/subword-nmt}}.

\paragraph{\textsc{nmt} Models} We use the base Transformer architecture \cite{attention} and include details on the exact architecture and training in Apendix~\ref{sec:sockeye}.

\paragraph{Selective Replacement} The divergence ranking models are trained using our public  implementation of divergentm\textsc{bert}~\cite{briakou-carpuat-2020-detecting}.\footnote{\url{https://github.com/Elbria/xling-SemDiv}} 
Synthetic divergences are generated starting from the $5{,}000$ top scoring WikiMatrix sentences based on \textsc{laser} score (i.e., seed equivalents). 
We fine-tune the “\textsc{bert}-Base Multilingual Cased” model~\cite{devlin-etal-2019-bert}
and set the margin equal to $5$ as per our original implementation. We use the same margin value for the margin score of Algorithm~\ref{alg:algo}. 
\footnote{Our divergentm\textsc{bert} yields $84$ F$1$ on a set of English-French human-annotated fine-grained divergences in WikiMatrix collected in our prior work~\cite{briakou-carpuat-2020-detecting}.}
%

\begin{table*}[!t]
    \centering
    \scalebox{0.8}{
    \begin{tabular}{l@{\hskip 0.8in}l@{\hskip 0.8in}l}
    \toprule[2pt]
    \addlinespace[0.5em]
    \textsc{\textbf{[el]}} & \textsc{ wikimatrix}  & Απεβίωσε στην Αθήνα στις 5 Ιουνίου 1979.\\
    & \hspace{0.5em} $\lfloor$ \hspace{0.3em} \color{gray}{\textsc{gloss}} & \it \color{gray}{He died in Athens on 5 June 1979.}\\
    \textsc{\textbf{[en]}} & \textsc{wikimatrix}  &  He died in \colorbox{yellow!30}{\hz London} on 5 June 1979.\\
    \textsc{\textbf{[en]}} &\textsc{synthetic translation}  & He died in \colorbox{green!20}{Athens} on 5 June 1979.\\
    \addlinespace[0.2em]
    \cmidrule{3-3}
    \addlinespace[0.2em]
    \textsc{\textbf{[el]}} & \textsc{wikimatrix}  & Ένας από τους οικισμούς που δημιούργησαν ήταν ο Καραβάς.\\
    & \hspace{0.5em} $\lfloor$ \hspace{0.3em} \color{gray}{\textsc{gloss}}  & \it \color{gray}{Karavas was one of the first settlements they created.} \\
    \textsc{\textbf{[en]}} & \textsc{wikimatrix}  & One of the first \colorbox{yellow!30}{towns} to be created was \colorbox{yellow!30}{Vila Barreto}.\\
    \textsc{\textbf{[en]}} & \textsc{synthetic translation}  & One of the first \colorbox{green!20}{settlements} to be created was \colorbox{green!20}{Karavas}.\\
    \addlinespace[0.2em]
    \cmidrule{3-3}
    \addlinespace[0.2em]
    \textsc{\textbf{[el]}} & \textsc{ wikimatrix}  & Και οι έξι λέβητες κατασκευάστηκαν από την Waagner-Biro. \\
    & \hspace{0.5em} $\lfloor$ \hspace{0.3em} \color{gray}{\textsc{gloss}}  & \it \color{gray}{All six boilers were manufactured by Waagner-Biro.}\\
    \textsc{\textbf{[en]}} & \textsc{wikimatrix}  &Boilers were \colorbox{yellow!30}{supplied} by Waagner-Biro.\\
    \textsc{\textbf{[en]}} & \textsc{synthetic translation}  & \colorbox{green!20}{All six} boilers were \colorbox{green!20}{manufactured} by Waagner-Biro.\\
    \addlinespace[0.2em]
    \cmidrule{3-3}
    \addlinespace[0.2em]
    \textsc{\textbf{[el]}} & \textsc{wikimatrix}  &Το Διδακτικό προσωπικό της Σχολής είναι υψηλού επιπέδου.\\
     & \hspace{0.5em} $\lfloor$ \hspace{0.3em} \color{gray}{\textsc{gloss}}  & \it \color{gray}{The school's teaching staff is of a high level.} \\
    \textsc{\textbf{[en]}} & \textsc{wikimatrix}  &The \colorbox{yellow!30}{medical research level} of the school is high.\\
    \textsc{\textbf{[en]}} & \textsc{synthetic translation}  &The \colorbox{green!20}{teaching staff} of the school is high.\\
    \addlinespace[0.2em]
    \cmidrule{3-3}
    \addlinespace[0.2em]
    \textsc{\textbf{[el]}} & \textsc{wikimatrix}  &Ανήκει στο τριπλό αστρικό σύστημα του Άλφα Κενταύρου. \\
    & \hspace{0.5em} $\lfloor$ \hspace{0.3em} \color{gray}{\textsc{gloss}}  & \it \color{gray}{It belongs to the Alpha Centauri triple star system.}\\ 
    \textsc{\textbf{[en]}} & \textsc{wikimatrix}  & \colorbox{yellow!30}{This is the} triple alpha process.\\
    \textsc{\textbf{[en]}} & \textsc{synthetic translation}  & \colorbox{green!20}{It belongs to the} triple star system of Alpha \colorbox{green!20}{Centauri}.\\
    \addlinespace[0.2em]
    \cmidrule{3-3}
    \addlinespace[0.2em]
   \textsc{\textbf{[el]}} & \textsc{ wikimatrix}  & Η εμφάνιση τυφώνων είναι σύνηθες φαινόμενο.\\ 
    & \hspace{0.5em} $\lfloor$ \hspace{0.3em}  \color{gray}{\textsc{gloss}} & \it \color{gray}{ The occurrence of hurricanes is a common phenomenon.}\\ 
    \textsc{\textbf{[en]}} & \textsc{wikimatrix}  & \colorbox{yellow!30}{It is extremely rare: There were only 10 known cases in 1998.} \\
    \textsc{\textbf{[en]}} & \textsc{synthetic translation}  & \colorbox{green!20}{The appearance of hurricanes is a common phenomenon.}\\
    \addlinespace[0.5em]
    \toprule[2pt]
    \end{tabular}}
    \caption{Randomly sampled WikiMatrix pairs with synthetic translations that satisfy $d>5$. 
    Selective replacement successfully revises divergences of different granularities (highlighted segments) in the original references.
    }
    
    \label{tab:examples_led}
\end{table*}

\section{Intrinsic Evaluation of Bitext Quality}

\subsection{Human evaluation}

We ask $3$ bilingual speakers to evaluate the quality of the \textsc{en}-\textsc{el} bitexts.
Given an original source sentence, they are asked to rank the original target and the candidate target in the order of their equivalence to the source. They are asked ``Which sentence conveys the meaning of the source better?'', and ties are allowed. A random sample of $100$ pairs from forward and backward \textsc{MT} is annotated.
 
As can be seen in Table~\ref{tab:intrinsic_evaluation_results}, $60\%$ of \textsc{all} synthetic candidates are better translations of the WikiMatrix reference, which confirms the potential of \nmt for improving over original translations. Further ablations confirm the benefits of selecting these synthetic candidates with the semantic equivalence condition. When the divergent scorer ranks a candidate higher than the original
by a small margin (i.e., $0\leq d \leq 5$ given $d=R(S_i,M_{S\rightarrow T}(T_i))-R(S_i,T_i))$), human evaluation shows that the candidate is actually better than the original only $51\%$ of the times.
When using our exact semantic equivalence condition~($d>5$), candidates are judged as more equivalent than the original $87.5\%$ of the times,
and annotations within this set have a stronger agreement (i.e., $0.688$ Kendall's τ). 
This indicates that
the condition 
$d>5$ identifies more clear-cut examples of synthetic translations that fix semantic divergences
in the original data and can be thus used for selective replacement of imperfect references by better quality translations.
%

\begin{table}[!t]
    \centering
    \scalebox{1.00}{
    \begin{tabular}{lcc}
    \rowcolor{gray!10}
    Candidate set &  \% Equivalized  & Kendall's τ   \\
    \addlinespace[0.5em]
    \textsc{all}   & $60.0\%$ & $0.321$  \\ 
    $d<0$ & $26.4\%$ & $0.157$ \\
    $0\leq d \leq 5$ & $51.0\%$ & $0.234$  \\
    \rowcolor{gray!5}
    $d>5$     & $\mathbf{87.5\%}$ &  $\mathbf{0.688}$\\
    \end{tabular}}
    \caption{Human evaluation results for all evaluated pairs and ablation sets
    for different thresholds on divergent score differences between candidates and originals~(i.e., $d$). 
    }\vspace{-0.5cm}
    \label{tab:intrinsic_evaluation_results}
\end{table}

\begin{table*}[!ht]
    \centering
    \scalebox{0.8}{
    \begin{tabular}{lrl@{\hskip 0.6in}r@{\hskip 0.6in}r@{\hskip 0.4in}r}
    \\\toprule[2pt]
    & & \textsc{\textbf{property}} & \textsc{\textbf{original}}  & \textsc{\textbf{revised}}  & $\delta$ \ph \\
    \cmidrule{4-6}
    \addlinespace[0.1cm]

   & $1:$ & \# Sentences   & $750{,}585$ & $750{,}585$          & $0.0\%$ \ph \\
   &  $2:$ & \# Tokens & $15{,}244{,}413$ & $15{,}239{,}474$ & {\color{burgundy!80}{$\mathbf{-0.3\%}$}} \da \\
    & $3:$ & \# Types & $358{,}681$ & $350{,}224$            & {\color{burgundy!80}{$\mathbf{-2.4\%}$}} \da \\
    & $4:$ & Average Length & $20.3$  & $20.3$                 & $0\%$ \ph \\
    & $5:$ & Average Coverage & $0.78$ & $0.83$                & {\color{britishracinggreen!80}{$\mathbf{+6.0\%}$}}   \ua   \\ 
    & $6:$ & \# \textsc{she/her/hers} Pronouns  & $45{,}028$ & $43{,}629$ & {\color{burgundy!80}{$\mathbf{-3.1\%}$}} \da \\
    \rot{\rlap{~\textit{English (\textsc{en})}}}  
    & $7:$ & \# \textsc{he/his/him} Pronouns & $185{,}356$ & $194{,}510$ & {\color{britishracinggreen!80}{$\mathbf{+4.7\%}$}} \ua\\
    & $8:$ & Complexity & $63.03$ & $53.61$                & {\color{burgundy!80}{$\mathbf{-14.9\%}$}} \da \\
    
    \addlinespace[0.1cm]
    \cmidrule{4-6}
    \addlinespace[0.1cm]

    & $9:$ & \# Sentences & $750{,}585$ & $750{,}585$             &  $0.0\%$ \ph \\
    & $10:$ & \# Tokens & $15{,}743{,}084$ & $15{,}611{,}937$  &  {\color{burgundy!80}{$\mathbf{-0.8\%}$}} \da\\
    & $11:$ & \# Types & $526{,}411$ & $519{,}558$             &  {\color{burgundy!80}{$\mathbf{-1.3\%}$}} \da\\
    & $12:$ & Average Length & $21.0$    & $20.8$                &  {\color{burgundy!80}{$\mathbf{-1.0\%}$}} \da \\
    & $13:$ & Average Coverage &  $0.77$       &     $0.83$      &  {\color{britishracinggreen!80}{$\mathbf{+7.0\%}$}} \ua \\
        \rot{\rlap{~\textit{Greek (\textsc{el})}}}  
    & $14:$ & \# {\small\textsc{Η/ΤΗΣ/ΤΗΝ}}  Pronouns &  $792{,}005$ & $776{,}947$      &  {\color{burgundy!80}{$\mathbf{-1.9\%}$}} \da\\
    & $15:$ & \# {\small{Ο/ΤΟΥ/ΤΟΝ}} Pronouns & $799{,}249$  & $794{,}275$      &  {\color{burgundy!80}{$\mathbf{-0.6\%}$}} \da \\
    & $16:$ & Complexity & $24.51$ & $17.85$                &  {\color{burgundy!80}{{$\mathbf{-27.0\%}$}}} \da
    \\\toprule[2pt]
    \end{tabular}}
    \caption{Comparison of original vs. revised bitext for \textsc{en-el}. $\delta$ gives percentage differences between them.}
    \label{tab:biases}
\end{table*}

Further inspection of the annotations reveals that most source-target WikiMatrix examples contain fine meaning differences ($56\%$). In those cases, we observe that most of the content between the sentences is shared, but either small segments are mistranslated (e.g., ``London'' instead of ``Athens'' in the first example of Table~\ref{tab:examples_led}), or some information is missing from either side of the pair (e.g., ``all six'' missing from the target side in the third example of Table~\ref{tab:examples_led}). Furthermore, more coarse-grained divergences are found less frequently ($12\%$)---in those cases, we notice that sentences are usually either topically related or structurally similar (e.g., length, syntax) with a few anchor words (e.g., last example in~Table~\ref{tab:examples_led}). Finally, $32\%$ of the times the original WikiMatrix pairs are perfect translations of each other.

%

\subsection{How do synthetic translations differ from originals?}

\begin{figure}[!t]
    \centering
    {\includegraphics[width=0.43\textwidth]{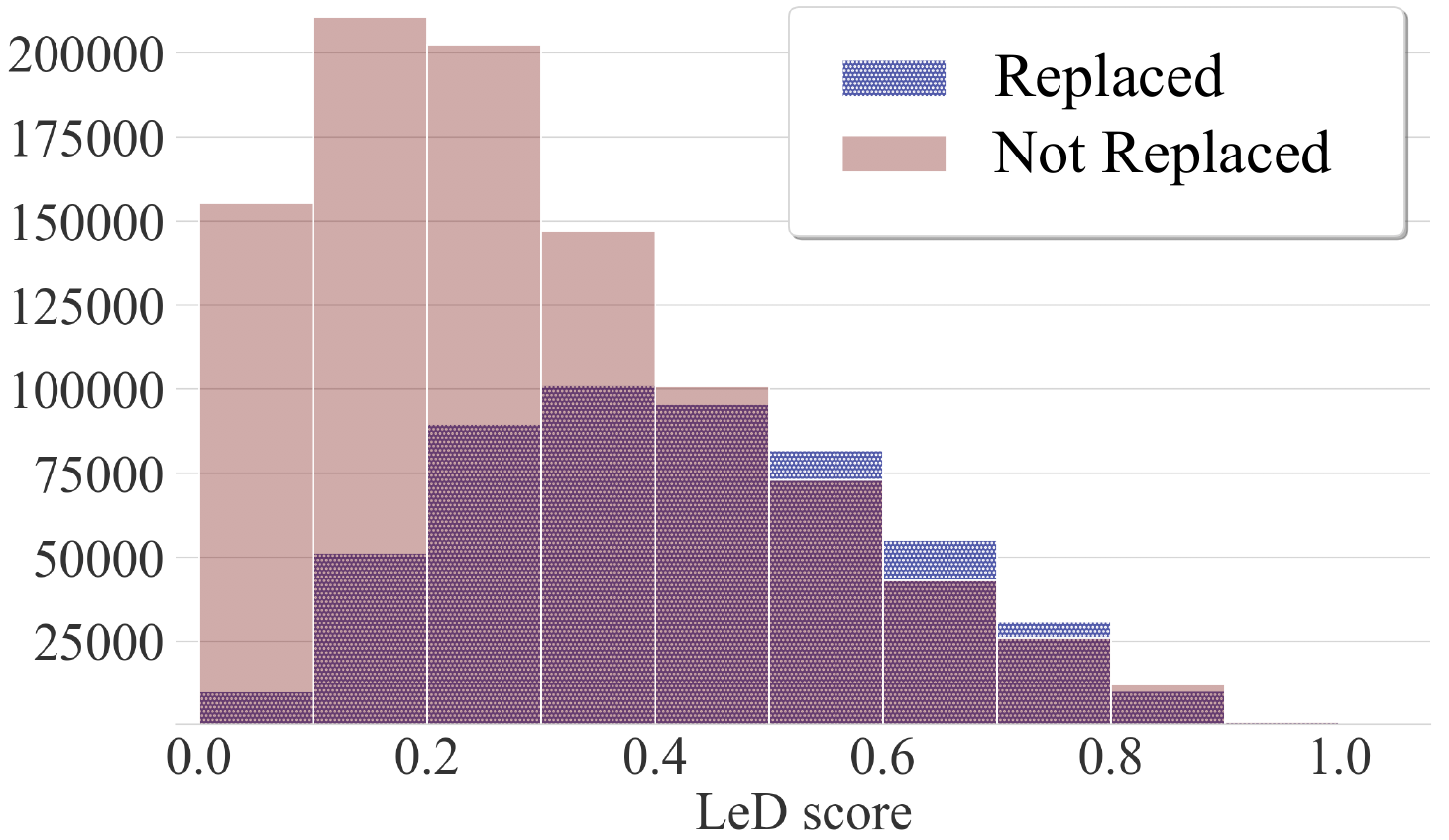}}
    \caption{LeD differences of original vs. synthetic translations (\textsc{el}$\rightarrow$\textsc{en}). 
    Replaced candidates share lexical content with the originals.
    }
    \label{fig:LeD_variants}
\end{figure}

Figure~\ref{fig:LeD_variants} presents the distribution of lexical differences (i.e., computed using LeD---a score that captures lexical differences based on the percentages of tokens that are not found in two sentences~\cite{Niu_Carpuat_2020}) between original and synthetic translations (in \textsc{en}) for candidates that replace and do not replace the originals.~\footnote{LeD details are in~Appendix \ref{sec:grammatical_gender_pronouns}.}
First, we observe that a substantial amount of synthetic translations that do not replace original references ($40\%$) corresponds to small \textsc{LeD} scores ($<0.1$), suggesting that the equivalence criterion could fall back to the original sentence not because of the poor quality  of candidate references, but rather due to them being already close to the originals.
Furthermore, all synthetic translated instances are represented in almost all bins, with fewer instances found on the extreme bins of $>0.7$ \textsc{LeD} scores. Finally, synthetic translations that replace original references are mostly concentrated within the range $[0.2, 0.6]$ of LeD scores. This indicates that they share lexical content with the original, which further supports the hypothesis that synthetic translations revise fine-grained meaning differences in WikiMatrix in addition to alignment noise.

%

\subsection{How does the revised bitext differ from the original?}\label{sec:revised_properties}

Table~\ref{tab:biases} presents differences in statistics of the original vs. revised  WikiMatrix \textsc{en-el} bitexts to shed more light on the impact of selectively using synthetic translation for bitext quality improvement.~\footnote{Details on the metrics are in~Appendix \ref{sec:grammatical_gender_pronouns}.}  The refined bitext exhibits higher coverage (i.e., ratio of source words being aligned by any target words; rows $5$ and $13$) and smaller complexity (i.e., the diversity of target word choices given a source word~\cite{chuntin}) compared to the original bitext. Moreover, the use of synthetic translations introduces small decreases in the lexical types covered in the final corpus~(i.e., rows $3$ and $11$), which is expected as the additional coverage in the original corpus might be a result of divergent texts. Those observations are in line with prior work that seeks to characterize the nature of synthetic translations used in other settings, such as knowledge distillation~\cite{chuntin, xu-etal-2021-distilled}.

While fixing divergent references contributes to this simplification effect, \nmt translations might also reinforce unwanted biases from the original bitext. For instance, the distribution of two grammatical gender pronouns on the English side is a little more imbalanced in the improved bitext than in the original (rows $6$-$7$ and $14$-$15$),
\footnote{We limit our analysis to \# occurrences for two grammatical gender pronouns. The complete list is in Appendix \ref{sec:grammatical_gender_pronouns}.}
likely due to gender bias in \nmt~\cite{stanovsky-etal-2019-evaluating}. This calls for techniques to mitigate such biases \citep{saunders-byrne-2020-reducing,stafanovics-etal-2020-mitigating} for \nmt and other downstream tasks.

\section{Extrinsic Evaluation of Bitext Quality}
Our previous analysis suggests that selective replacement of divergent references with synthetic translations results in bitext of \textit{improved quality}, with reduced level of noises and easier word-level mappings between the two languages, when compared to the original WikiMatrix corpus. To better understand how those differences impact downstream tasks, we contrast the improved bitext with the original through a series of extrinsic evaluations for \textsc{en-el} and \textsc{en-ro} languages that rely on parallel texts as training samples~(see \S\ref{sec:downstream_tasks}). First, we focus on the recent state-of-the-art unsupervised \textsc{bli} approach of \citet{shi-etal-2021-bilingual} that relies on word-alignments of extracted bitexts. 
Second, we follow the recent bitext quality evaluation frameworks adopted by the ``Shared Task on Parallel Corpus Filtering and Alignment''  \cite{koehn-etal-2020-findings} and built neural machine translation systems from scratch and by continued training on a multilingual pre-trained transformer model. 
Finally, we conduct extensive ablation experiments to test the impact of using synthetic translations without the semantic equivalence condition and contrast with familiar techniques used by prior work~(see \S\ref{sec:ablations}). 

%
%
\begin{table*}[!t]
    \centering
    \scalebox{1.00}{
    \begin{tabular}{l@{\hskip 0.1in}l@{\hskip 0.4in}lllr@{\hskip 0.6in}omh}
    \toprule[2pt]
    &       &    \multicolumn{4}{c}{\textit{All}} & \textit{Low} & \textit{Medium} & \textit{High} \\
     \textbf{\textsc{pair}}   & \textbf{\textsc{bitext}} & \multicolumn{1}{c}{Precision} & Recall & F1 & \textsc{oov} rate  & \multicolumn{3}{c}{Precision}  \\\toprule
        \multirow{2}{*}{\rotatebox{0}{\textsc{el-en \Big\{}}} & Original     & $76.2$\phh       & $58.1$ & $65.9$ & $\mathbf{6.7\%}$ &  $59.4$ & $76.6$ & $81.4$\\
                                                        & Revised  & $\mathbf{77.6^*}$ & $\mathbf{58.6^*}$ & $\mathbf{66.8^*}$ & $7.5\%$ & $\mathbf{60.4^*}$ & $\mathbf{78.4^*}$ & $81.6$\\

       \multirow{2}{*}{\rotatebox{0}{\textsc{en-ro} \Big\{}}  & Original     & $89.2$ & $69.4$ & $78.1$ & $\mathbf{15.8\%}$ &  $78.6$ & $86.9$ & $87.1$\\
                                                        & Revised  & $\mathbf{90.8^*}$ &  $\mathbf{71.3^*}$ & $\mathbf{79.8^*}$ & $16.5\%$ & $\mathbf{80.0^*}$  & $\mathbf{87.5^*}$ & $86.9$\\
                                \toprule[2pt]
    \end{tabular}}
    \caption{
    Unsupervised \textsc{bli} extrinsic evaluation results on \textsc{muse} for the entire dataset (\textit{All}) and on subsets binned by frequency (i.e., right-most highlighted columns).
    Revised bitexts yield statistically significant~($*$) improvements over the original bitexts overall and for low-to-medium frequency dictionary entries.}
    \label{tab:BLI_main}
\end{table*}

\subsection{Experimental Set-Up}

\paragraph{\textsc{bli}} 
The task of \textsc{bli} aims to induce a bilingual lexicon consisting of word translations in two languages. 
We experiment with the recently proposed method of \citet{shi-etal-2021-bilingual} that combines extracted bitext and unsupervised word alignment to perform fully unsupervised induction based on extracted statistics of aligned word pairs.
The induced lexicons are evaluated based on \textsc{muse}~\cite{lample2018word} consisting of $45{,}515$ and $80{,}815$ dictionary entries for \textsc{el-en} and \textsc{en-ro}, respectively.\footnote{\url{https://github.com/facebookresearch/MUSE}} 
We extract word alignments using m\textsc{bert}-based \textit{Simalign}\footnote{\url{https://github.com/cisnlp/simalign}} \cite{jalili-sabet-etal-2020-simalign} and statistics based on the implementation of \citet{shi-etal-2021-bilingual}.\footnote{\url{https://github.com/facebookresearch/bitext-lexind}}

\paragraph{\textsc{mt}} 
We experiment with \textsc{mt} tasks following two approaches:
(1) training standard transformer seq2seq models from scratch;
(2) continued training for mT5~\cite{xue-etal-2021-mt5}, a multilingual pre-trained text-to-text transformer. 
We evaluate translation quality with \bleu~\cite{papineni-etal-2002-bleu}\footnote{\url{https://github.com/mjpost/sacrebleu}} on the official development and test splits of the \ted corpus~\cite{qi-etal-2018-pre}. \footnote{Data statistics are found in Appendix~\ref{sec:data_statistics_appendix}.} 
For (1) we follow the experimental settings described in~\S\ref{sec:setup_human}. 
For (2) we initialize the weights of transformer with ``mT5-small'' which consists of $300$M parameters,\footnote{\url{https://github.com/google-research/multilingual-t5}}. We use the \texttt{simpletransformers} implementation.\footnote{\url{https://github.com/ThilinaRajapakse/simpletransformers}} We fine-tune for up to $5$ epochs and include the parameter settings in Appendix~\ref{sec:mt5_settings}.

\paragraph{Ablation Settings} 
We compare the \nmt models trained on the variants of the synthetic bitext to isolate the impact of replacement criteria and different candidates.\footnote{Results on development sets are in Appendix \ref{sec:dev_results}.}
For the former, we experiment with the \textbf{rejuvenation} approach of \citet{jiao-etal-2020-data} that replaces original references with forward translated candidates for the $10\%$ least active original samples measured by \nmt probability scores.
Moreover, we experiment with \textbf{forward} and \textbf{backtranslation} baselines trained on bitexts that consist solely from target- or source-side candidate sentences (i.e., original references are entirely excluded) and with ablations that consider either forward or backward candidates for the proposed semantic equivalence condition.
Finally, we consider two alternatives to 
the \textbf{semantic equivalence} condition based on divergent scores: the
\textbf{ranking} condition replaces a candidate if it scores higher than the original (i.e., margin with $d=0$) and the \textbf{thresholding} condition adds the additional constraint that candidates should rank higher than a threshold to replace the original pair.

\subsection{Extrinsic Evaluation Results}\label{sec:downstream_tasks}

\paragraph{\textsc{bli}}
Table~\ref{tab:BLI_main} presents results for unsupervised \textsc{bli} on the \textsc{muse} gold-standard dictionaries, for \textsc{el-en} and \textsc{en-ro}. Across languages, the revised bitexts induce better lexicons compared to the original WikiMatrix. Crucially, improvements are reported both in terms of Recall---which connects to the observation that the revised bitext exhibits higher coverage than the original and in terms of Precision---which connects to the noise reduction effect that impacts the extracted word alignments. Additionally, a break-down on the Precision of the induced lexicons binned by the frequency of \textsc{muse} source-side entries (i.e., last $3$ columns in Table~\ref{tab:BLI_main}) reveals that the improvements come from better induction of low- and medium-frequency words, which we expect are more sensitive to noisy misalignments that result from divergent bitext. Finally, those improvements are reported despite the small increase of the \textsc{oov} rate in the revised lexicons that results from the decrease in the lexical types covered in it, as mentioned in the analysis (i.e., ~\S\ref{sec:revised_properties}).

Furthermore, following the advice of \citet{kementchedjhieva-etal-2019-lost} who raise concerns on \textsc{bli} evaluations based on gold-standard pre-defined dictionaries, we accompany our evaluation with manual verification to confirm that our conclusions are consistent with those of the automatic evaluation. Concretely, we manually check the \textit{false positives} induced translation pairs from the original vs. the improved bitext. We found that $65/80$ are \textit{false false positives} (due to incompleteness of pre-defined dictionaries) for the improved bitext and $51/80$ for the original (see Appendix \ref{sec:bli_false_false_positives} for the complete list). This confirms that the metric improvements we observe are meaningful and suggests that the improved bitext help learn better mappings between source and target words. 

\paragraph{\textsc{mt}}

\begin{table}[!t]
    \centering
     \scalebox{1.0}{
    \begin{tabular}{l@{\hskip 0.35in}c@{\hskip 0.35in}c}
    \toprule[2pt]
    \textsc{\textbf{pair}} & \textbf{\textsc{original}} & \textbf{\textsc{revised}} \\
    \toprule[0.5pt]
    \textsc{el}$\rightarrow$\textsc{en} & $28.15$ {\small$\pm 0.13$} & $\mathbf{29.63}$ {\small $\mathbf{\pm 0.29}$} \\
    \textsc{en}$\rightarrow$\textsc{el} & $27.08$ {\small$\pm 0.18$} & $\mathbf{27.89}$ {\small$\mathbf{\pm 0.05}$} \\
    \textsc{ro}$\rightarrow$\textsc{en} & $23.68$ {\small$\pm 0.12$} & $\mathbf{24.54}$ {\small$\mathbf{\pm 0.06}$} \\
    \textsc{en}$\rightarrow$\textsc{ro} & $20.65$ {\small$\pm 0.10$} & $\mathbf{20.84}$ {\small$\mathbf{\pm 0.04}$} \\
    \toprule[2pt]
    \end{tabular}}
    \caption{\textsc{bleu} on \textsc{nmt} training from scratch.
    }
    \label{tab:mt}
\end{table}
\begin{figure}[!t]
    \centering
    \includegraphics[scale=0.4]{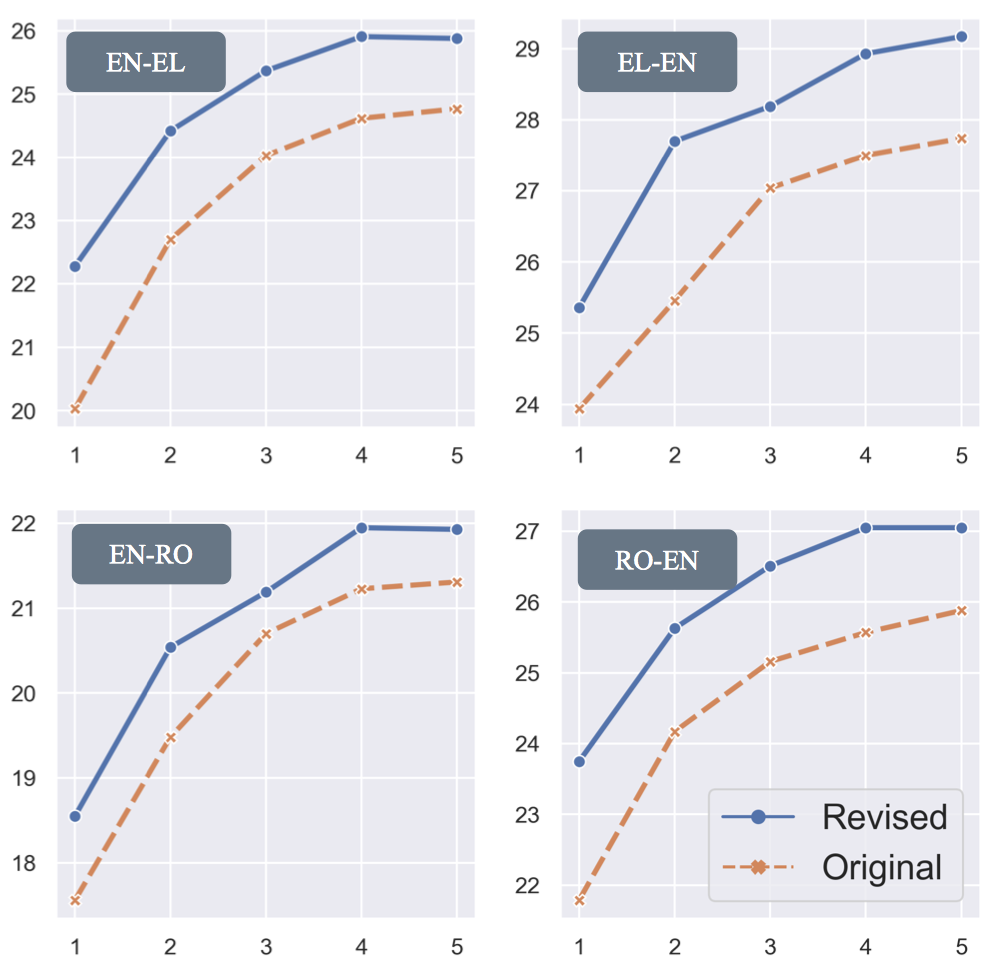}
    \caption{\textsc{bleu} scores across epochs (x-axis) for continued training on mt5. The revised bitext improves translation quality compared to the original for all epochs and translation tasks.}\vspace{-0.3cm}
    \label{fig:mt5}
\end{figure}
\begin{table*}[!t]
    \centering
     \scalebox{0.73}{
    \begin{tabular}{rl@{\hskip 0.5in}l@{\hskip 0.5in}cc@{\hskip 0.5in}rrrc}
    
    & \textbf{\textsc{selective}} & \textbf{\textsc{data}} & & & \multicolumn{4}{c}{\textsc{\textbf{Bitext statistics}}} \\
    & \textbf{\textsc{replacement}} & \textbf{\textsc{types}}  & \textsc{bleu} & $\delta$ & \dto &  \dtf &  \dtb & \textbf{\textsc{vis.}}\\
    
    \rowcolor{gray!10}
    \multicolumn{9}{c}{\textbf{\textsc{en}}$\rightarrow$\textbf{\textsc{el}}}\\
    \addlinespace[0.5em]
    
    $1:$ & \xmark    & \dto               &   $27.08$ {\small $\pm 0.18$} & $-$      & $100\%$ & $0\%$   & $0\%$   & \dataviz{1}{0}{0}   \\
    $2:$ &\xmark    & \dtf               &   $27.45$ {\small$\pm 0.06$} & $+0.36$  & $0\%$   & $100\%$ & $0\%$   & \dataviz{0}{1}{0}    \\
    $3:$ &\xmark    & \dtb               &   $26.22$ {\small$\pm 0.26$} & $-0.86$  & $0\%$   & $0\%$   & $100\%$ & \dataviz{0}{0}{1}    \\
    $4:$ &Rejuvenation & \dto \dtf           &   $27.24$ {\small$\pm 0.11$} & $+0.16$  & $90\%$  & $10\%$  & $0\%$   & \dataviz{0.9}{0.1}{0} \\

    $5:$ &Ranking         & \dto \dtf         &   $27.21$ {\small$\pm 0.43$} & $+0.13$  & $22\%$  & $78\%$ & $0\%$   & \dataviz{0.22}{0.78}{0}    \\
    $6:$ & Thresholding    & \dto \dtf        &   $27.56$ {\small$\pm 0.11$} & $+0.48$  & $78\%$  & $21\%$ & $0\%$   & \dataviz{0.78}{0.21}{0}    \\
    $7:$ & Semantic equivalence  & \dto \dtf         &   $27.64$ {\small$\pm 0.22$} & $+0.56$  & $63\%$  & $37\%$ & $0\%$   & \dataviz{0.63}{0.37}{0}   \\
    $8:$ & Semantic equivalence         & \dto \dtb         &   $27.61$ {\small$\pm 0.09$} & $+0.52$  & $66\%$  & $0\%$  & $34\%$  & \dataviz{0.66}{0}{0.34}  \\ 
    $9:$ & Semantic equivalence           & \dto \dtf \dtb      &   $\mathbf{27.89}$ {\small$\pm 0.05$} & $\mathbf{+0.81}$   & $50\%$ & $23\%$ & $27\%$  &\dataviz{0.50}{0.23}{0.27} \\

    \addlinespace[0.5em]
    \rowcolor{gray!10}
    \multicolumn{9}{c}{\textbf{\textsc{el}}$\rightarrow$\textbf{\textsc{en}}}\\
    \addlinespace[0.5em]

    $10:$ & \xmark    & \dto                      & $28.15$ {\small$\pm0.13$} & $-$     &  $100\%$ & $0\%$ & $0\%$  &  \dataviz{1}{0}{0} \\ 
    $11:$ & \xmark    & \dtf        & $28.16$ {\small$\pm 0.17$} & $+0.01$ &  $0\%$   & $100\%$ & $0\%$ &  \dataviz{0}{1}{0} \\
    $12:$ & \xmark    & \dtb               & $28.38$ {\small$\pm0.09$} & $+0.23$ &  $0\%$   & $0\%$ & $100\%$ & \dataviz{0}{0}{1} \\ 
    $13:$ & Rejuvenation &  \dto \dtf                  & $28.27$ {\small$\pm0.12$} & $+0.12$ &  $90\%$  & $10\%$ & $0\%$ &  \dataviz{0.9}{0.1}{0}\\
    
    $14:$ & Ranking         & \dto \dtf  & $28.81$ {\small$\pm0.13$} & $+0.67$ &  $26\%$ & $74\%$ & $0\%$ &  \dataviz{0.26}{0.74}{0}   \\
    $15:$ & Thresholding    & \dto \dtf  & $28.79$ {\small$\pm0.17$} & $+0.64$ &  $81\%$ & $19\%$ & $0\%$ &  \dataviz{0.81}{0.19}{0} \\
    $16:$ & Semantic equivalence         & \dto \dtf  & $29.00$ {\small$\pm0.15$} & $+0.85$ &  $66\%$ & $34\%$ & $0\%$ &  \dataviz{0.66}{0.34}{0}  \\ 
    $17:$ & Semantic equivalence          & \dto \dtb   & $29.19$ {\small$\pm0.25$} & $+1.05$ & $63\%$ & $0\%$ & $37\%$ &  \dataviz{0.63}{0}{0.37} \\
    
    $18:$ & Semantic equivalence         & \dto \dtf \dtb & $\mathbf{29.63}$ {\small $\pm 0.29$} & $\mathbf{+1.49}$ &   $50\%$   & $27\%$ & $23\%$ & \dataviz{0.5}{0.27}{0.23} \\

    \end{tabular}}
    \caption{\textsc{bleu} results (averages of $3$ seeds) on \textsc{en}$\leftrightarrow$\textsc{el} \nmt.  $\delta$ denotes average improvements over the original bitext. Bitext statistics give percentage of original (\dto), forward (\dtf), and backward (\dtb) translated candidates. First column shows the selective replacement condition for candidate replacement (when applicable).}\vspace{-0.3cm}
    \label{tab:en2el_bleu}
\end{table*}

Table~\ref{tab:mt} presents translation quality (\bleu) on \textsc{en}$\leftrightarrow$\textsc{ro} and \textsc{en}$\leftrightarrow$\textsc{el} tasks for \textsc{mt} training from scratch and Figure~\ref{fig:mt5} shows translation quality of mT5 continued training across epochs. Across tasks and settings, the revised bitext yields better translation quality than the original WikiMatrix data. 
The consistent improvements we observe across the two settings suggest that the properties of the synthetic translations that replace original samples and bring those improvements are invariant to specific models. Moreover,
the magnitude of improvements is larger in the continued training setting compared to training from scratch (e.g., $\sim+0.8$ vs. $\sim+1.5$, for \textsc{en}$\rightarrow$\textsc{el}; $ \sim+0.2$ vs. $\sim+1.5$, for \textsc{ro}$\rightarrow$\textsc{en}). The latter suggests that improvements from using synthetic samples do not only come from the normalization effect (i.e., synthetic samples are easier to model by \textsc{nmt}) but also connect to the reduced noise in the training samples. This further complements our hypothesis that synthetic translations can improve the quality of imperfect references that should, in principle, yield noisy training signals---and thus impact the resulting quality---of different \textsc{mt} models.


%
%
\subsection{Ablation Study}\label{sec:ablations}
Table~\ref{tab:en2el_bleu} compares the translation quality (\textsc{bleu}) of \textsc{nmt} systems trained on different synthetic translations. By forcing the semantic equivalence condition when deciding whether a synthetic translation replaces an original, we revise $50\%$ of the latter
yielding the best results across directions with significant improvements~(i.e, increases do not lie within $1$ stdev of the original's bitext performance) of $+0.81$~(\textsc{en}$\rightarrow$\textsc{el}, row $9$) and $+1.49$~(\textsc{el}$\rightarrow$\textsc{en}, row $18$)
points over the original bitext.

\paragraph{Impact of semantic equivalence condition} Table~\ref{tab:en2el_bleu} shows that naively disregarding the original references and training only on synthetic translations gives mixed results: training on \textit{forward-translated} references only~(i.e., row $2$) gives small improvements ($+0.36$) over the model trained on WikiMatrix for \textsc{en}$\rightarrow$\textsc{el}, while it performs comparably to it for \textsc{el}$\rightarrow$\textsc{en}~(i.e.,~row~$11$). On the other hand, training on \textit{backward} data only~(i.e.,~row~$12$) improves \textsc{bleu} by a small margin ($+0.23$) for \textsc{mt} into \textsc{en} while it hurts \textsc{bleu} when translating into \textsc{el}~(i.e.,~row~$3$). This indicates that the good quality of the synthetic translations  cannot be taken for granted and motivates replacing original pairs under conditions that account for semantic controls. 

The latter is further confirmed by results on the rejuvenation baseline: replacing candidates for the $10\%$ of the most inactive WikiMatrix samples results in small and insignificant
increases in \textsc{bleu} when compared to models trained on original WikiMatrix data~(i.e., rows $1$-$4$ and $10$-$13$).  This indicates that rejuvenation might not be well-suited to lower resource settings than the ones it was originally tested on \citep{jiao-etal-2020-data}. 
The rejuvenation technique might be affected by the decreased \nmt quality and calibration in lower resource settings. By contrast, using synthetic translations with semantic control mitigates their impact. 

Finally, all three semantic control variants based on divergent scores
yield bitexts that improve \textsc{bleu} compared to the original WikiMatrix~(i.e.,~rows~$5$-$8$~and~$14$-$18$). Among them, 
the \textit{margin} condition is the most successful, followed by the \textit{thresholding} variant. The breakdown of training statistics reveals the reason behind their differences:  the \textit{thresholding} condition is a more strict constraint as it only allows synthetic candidates to replace the original pairs if they are predicted as exact equivalents, allowing for fewer revisions of divergent pairs in WikiMatrix. By contrast, the condition based on \textit{margin} is a contrastive approach that allows for more revisions of the original data (i.e., a candidate might be a more fine-grained divergent of the source). The \textit{ranking} criterion is the least successful method---this is expected as the divergence ranker is not trained as a regression model. 

\paragraph{Impact of bi-directional candidates} 
Considering both forward (\dtf ) and backward (\dtb) translated candidates during selective replacement yields to further improvements ($0.22$-$0.44$ points) over bitext induced by the semantic equivalence condition with candidates from a single \nmt model~(i.e.,~rows~$7$-$9$ and $16$-$18$). When forward and backward candidates are considered independently, they replace $34-37\%$ of the original pairs; in contrast, when considered together, they replace $50\%$ of original WikiMatrix pairs. As a result, there is no perfect overlap between the original pairs replaced by the forward vs. backward model, which  motivates the use of both to revise more divergences in WikiMatrix. This finding raises the question of whether using synthetic translations from both directions might benefit other scenarios, such as knowledge distillation.

\section{Conclusion}

This paper explored how synthetic translations can be used to revise bitext, using \nmt models trained on the exact same data we seek to revise. Our extensive empirical study surprisingly shows that, even without access to further bilingual data or supervision, this approach improves the quality of the original bitext, especially when synthetic translations are generated in both translation directions and selectively replace the original using a semantic equivalence criterion. Specifically, our intrinsic evaluation showed that synthetic translations are of sufficient quality to improve over the original references, in addition to ``normalizing'' the bitext as suggested by prior work and corpus level statistics~\citep{chuntin, xu-etal-2021-distilled}.
Extrinsic evaluations further show that the replaced synthetic translations provide more useful signals for \textsc{bli} tasks and \nmt training in two settings (i.e., training from scratch and continued training). 

These findings provide a foundation for further exploration of the use of synthetic bitext. First, we focused our empirical study on language pairs and datasets where revising bitexts is the most needed and most likely to be useful: the resources available for these languages are not so large that mined bitext can simply be ignored or filtered with simple heuristics, yet there is enough data to build \nmt systems of reasonable quality (i.e., $\sim600$K segments for \textsc{en-ro}, and $\sim750$K for \textsc{en-el}). While in principle, selective replacement of divergent references with synthetic translations should port to high-resource settings, where \nmt is as good or better than for the languages considered in this work, other techniques are likely needed in low-resource settings where \nmt quality is too low to provide reliable candidate translations. Second, having established that the revised bitext improves the quality of the original bitext in isolation, it remains to be seen how to best revise bitexts in more heterogeneous scenarios with diverse sources of parallel or monolingual corpora. Overall, as synthetic data generated by \nmt is increasingly used to improve cross-lingual transfer in multilingual NLP, our study motivates taking a closer look at the properties of synthetic samples to better understand how they might impact downstream tasks beyond raw performance metrics. All bitexts are available at: \url{https://github.com/Elbria/xling-SemDiv-Equivalize}.

\section*{Acknowledgements}

We thank Marjan Ghazvininejad, Luke Zettlemoyer, Sida Wang, 
Sweta Agrawal, Jordan Boyd-Graber, Pedro Rodriguez,
the anonymous reviewers and the \textsc{clip} lab at \textsc{umd} for helpful comments. 
This material is based upon work supported by the National Science Foundation under Award No.\ $1750695$. Any opinions, findings, and conclusions or recommendations expressed in this material are those of the authors and do not necessarily reflect the views of the National Science Foundation.

\bibliography{anthology,custom}
\bibliographystyle{acl_natbib}
\clearpage
\appendix

\section{Details on bitext analysis}\label{sec:grammatical_gender_pronouns}
\paragraph{Complexity} We follow~\citet{chuntin} and compute the 
corpus complexity as a measure of translation uncertainty. Concretely, having access to an alignment model (here, \texttt{fast-align}), the complexity of a corpus $d$ is computed by averaging the entropy of target words $y$ conditioned on the source $x$, 
$L(d)=\frac{1}{|\mathcal{V}_x|}\sum_{x \in \mathcal{V}_x} H(y|x)$.
\paragraph{Coverage} We follow~\citet{tu-etal-2016-modeling} and measure the coverage of each source-target parallel pair as the ratio of source words being aligned to target words, having access to an alignment model (here, \texttt{fast-align}). We compute the coverage for source-target and target-source bitexts separately. Corpus-level statistics correspond to average sentence-level results.
\paragraph{Grammatical Gender Pronouns}
The complete lists of grammatic gender pronouns we use for \textsc{el} are:
[ο, του, τον, αυτός,  αυτού,  αυτόν, εκέινος, εκέινου, εκείνον, οποίος, οποίου, οποίον  ] and [η, της, την, αυτήν,  αυτής,  αυτήν, εκέινη, εκέινης, εκείνην, οποία, οποίας, οποίαν].
\paragraph{Lexical Differences (LeD)} We follow~\cite{Niu_Carpuat_2020} and compute the Lexical Differences score between two sentences $S_1$ and $S_2$ as the percentage of tokens that are not found in both, LeD$=\frac{
1}{2}(\frac{|S_1 /\ S_2|}{|S_1|}) + \frac{|S_2 /\ S_1|}{|S_2|}$.

\section{Result on development sets}\label{sec:dev_results}
Table~\ref{tab:dev_results} presents results on the main and secondary \textsc{nmt} tasks on \textsc{ted} developments sets. The refined bitext leads to consistent and significant improvements in \bleu across language-pairs and translation directions. 
\begin{table}[!t]
    \centering
    \scalebox{1.0}{
    \begin{tabular}{rccc}
    
    \rowcolor{gray!10}
    \multicolumn{4}{c}{\textit{Table}~\ref{tab:en2el_bleu}} \\
    \rowcolor{gray!5}
   \multicolumn{2}{c}{\textsc{en}$\rightarrow$\textsc{el}} & \multicolumn{2}{c}{\textsc{el}$\rightarrow$\textsc{en}}\\
    $1:$     & $25.50 \pm 0.15$ & $10:$ & $27.98 \pm 0.18$\\
    $2:$     & $25.52 \pm 0.07$ & $11:$ & $27.92 \pm 0.15$\\
    $3:$     & $24.55 \pm 0.25$ & $12:$ & $27.70 \pm 0.15$\\
    $4:$     & $25.35 \pm 0.14$ & $13:$ & $27.99 \pm 0.15$\\
    $5:$     &  $25.27 \pm 0.41$ & $14:$ & $28.36 \pm0.13$*\\
    $6:$     & $25.66 \pm 0.05$* & $15:$ & $28.34 \pm0.18$*\\
    $7:$     & $25.73 \pm 0.14$* &  $16:$ & $28.66 \pm0.14$* \\
    $8:$     & $25.71 \pm 0.19$* & $17:$ & $28.65 \pm0.27$*\\
    $9:$     & $\mathbf{25.91 \pm 0.09}$* & $18:$ & $\mathbf{29.00 \pm0.26}$*\\
    
    \addlinespace[1em]
    
    \rowcolor{gray!10}
    \multicolumn{4}{c}{\textit{Table}~\ref{tab:mt}} \\
    \rowcolor{gray!5}
    \multicolumn{2}{c}{\textsc{en}$\rightarrow$\textsc{ro}} & \multicolumn{2}{c}{\textsc{ro}$\rightarrow$\textsc{en}}\\
    
    $1:$ & $21.94\pm0.11$ & $3:$  & $24.98\pm0.16$ \\
    $2:$ & $\mathbf{22.05\pm0.03}$* & $4:$ & $\mathbf{26.11\pm0.20}$* \\
    \end{tabular}}
    \caption{\bleu results on the \textsc{ted} developments sets for each of the results of Tables~\ref{tab:en2el_bleu} and \ref{tab:mt} (enumeration follows the main text Tables). * denotes one standard deviation improvements over the original bitexts.}
    \label{tab:dev_results}
\end{table}


\section{\texttt{Sockeye2} configuration details}\label{sec:sockeye}

We use the base Transformer architecture \cite{attention}.
with embedding size of $512$, transformer hidden size of $2{,}048$, $8$ attention heads, $6$ transformer layers, and dropout of $0.1$.
Target embeddings are tied with the output layer weights. 
We train with label smoothing ($0.1$). We optimize with Adam~\cite{Kingma2015AdamAM} with a batch size of $4{,}096$ tokens and checkpoint models every $1{,}000$ updates. The initial learning rate is $0.0002$, and it is reduced by $30$\% after $4$ checkpoints without validation perplexity improvement. We stop training after $20$ checkpoints without improvement. We select the best checkpoint based on validation \bleu~\cite{papineni-etal-2002-bleu}. All models are trained on a single GeForce \textsc{gtx} $1080$ \textsc{gpu}. Tables~\ref{tab:sockeye_main_config} presents details of \nmt training with~\texttt{Sockeye2}.
\begin{table}[!t]
    \centering
    \scalebox{0.84}{
    \begin{tabular}{|l|}
    \hline
    \texttt{--weight-tying-type="trg\_softmax"} \#uni-\textsc{nmt}\\
    \texttt{--weight-tying-type="src\_trg\_softmax"} \#bi-\textsc{nmt} \\
    \texttt{--num-words $5000$:$5000$} \\
    \texttt{--label-smoothing $0.1$} \\
    \texttt{--encoder transformer} \\
    \texttt{--decoder transformer} \\
    \texttt{--num-layers $6$} \\
    \texttt{--transformer-attention-heads $84$} \\
    \texttt{--transformer-model-size $512$} \\
    \texttt{--num-embed $512$} \\
    \texttt{--transformer-feed-forward-num-hidden $2048$} \\
    \texttt{--transformer-preprocess n} \\
    \texttt{--transformer-postprocess dr} \\
    \texttt{--gradient-clipping-type none} \\
    \texttt{--transformer-dropout-attention $0.1$} \\
    \texttt{--transformer-dropout-act $0.1$} \\
    \texttt{--transformer-dropout-prepost $0.1$} \\
    \texttt{--max-seq-len $80$:$80$}\\
    \texttt{--batch-type word} \\
    \texttt{--batch-size $2048$} \\
    \texttt{--min-num-epochs $3$} \\
    \texttt{--initial-learning-rate $0.0002$} \\
    \texttt{--learning-rate-reduce-factor $0.7$} \\
    \texttt{--learning-rate-reduce-num-not-improved $4$} \\
    \texttt{--checkpoint-interval $1000$} \\
    \texttt{--keep-last-params $30$} \\
    \texttt{--max-num-checkpoint-not-improved $20$} \\
    \texttt{--decode-and-evaluate $1000$} \\
    \hline
    \end{tabular}}
    \caption{\nmt configurations on \texttt{Sockeye2}}
    \label{tab:sockeye_main_config}
\end{table}
%
\newpage
\section{mt5 configuration details}\label{sec:mt5_settings}
Tables~\ref{tab:mT5_main_config} presents details of continued training of mT5 on \texttt{SimpleTransformers}.
\begin{table}[!t]
    \centering
    \scalebox{0.84}{
    \begin{tabular}{|l|}
    \hline
    \texttt{max-seq-length $100$} \\
    \texttt{train-batch-size $10$} \\
    \texttt{eval-batch-size $10$} \\
    \texttt{num-train-epochs $5$} \\
    \texttt{scheduler 'cosine schedule with warmup'} \\
    \texttt{evaluate-during-training True} \\
    \texttt{evaluate-during-training-steps $10000$} \\
    \texttt{learning-rate $0.0003$} \\
    \texttt{optimizer 'Adafactor'} \\
    \texttt{use-multiprocessing False} \\
    \texttt{save-model-every-epoch True} \\
    \texttt{use-early-stopping False} \\
    \texttt{do-lower-case True} \\
    \hline
    \end{tabular}}
    \caption{\nmt configurations for continued training of mT5 on \texttt{SimpleTransformers}.}
    \label{tab:mT5_main_config}
\end{table}

\section{Data Statistics}\label{sec:data_statistics_appendix} Table~\ref{tab:data_stats} presents data statistics for WikiMatrix training data, and \textsc{ted} evaluation sets.
\begin{table}[!t]
    \centering
    \scalebox{0.9}{
    \begin{tabular}{lllr}
    \toprule
    \textbf{\textsc{language pair}} &  \textbf{\textsc{training}} & \textbf{\textsc{dev.}} & \textbf{\textsc{Test}} \\
    \bottomrule
    \addlinespace[0.2em]
    \textsc{el-en} & $750{,}585$  & $3{,}344$ & $4{,}431$\\
    \textsc{ro-en} & $582{,}134$  & $3{,}904$ & $4{,}631$ \\
    \bottomrule
    \end{tabular}}
    \caption{Data statistics after pre-processing.}
    \label{tab:data_stats}
\end{table}

\begin{table}[!t]
    \centering
    \scalebox{0.9}{
    \begin{tabular}{lll}
    \toprule
    \textbf{\textsc{language pair}} & \multicolumn{1}{c}{\textsc{uni-nmt}} & \multicolumn{1}{c}{\textsc{bi-nmt}}\\
    \bottomrule
    \addlinespace[0.2em]
    \textit{\en} $\rightarrow$ \textit{\el}     & $27.89\pm0.29$ & $27.92\pm0.06$ \\ 
     \textit{\el} $\rightarrow$ \textit{\en}        & $29.63\pm0.29$ & $29.57\pm0.36$\\ 
    \textsc{ro} $\rightarrow$ \textit{\en}       &  $24.54\pm0.06$        & $24.69\pm0.11$\\ 
    \textsc{en} $\rightarrow$ \textsc{ro}       & $20.84\pm0.04$         & $20.73\pm0.12$\\ 
    \bottomrule
    \end{tabular}}
    \caption{\bleu scores for \textsc{nmt} on equivalized bitexts using uni- (\textsc{uni-nmt})  vs. bi-directional \nmt models (\textsc{bi-nmt}). Equivalizing the bitext with \textsc{bi-nmt} \textsc{nmt} yields comparable \bleu  with \textsc{uni-nmt}.
    }
    \label{tab:speeding_up}
\end{table}%

\section{Manual inspection of \textsc{bli}}\label{sec:bli_false_false_positives}
Table~\ref{tab:ground_truth_bli} presents manual analysis results on False Positives entries of the \textsc{muse} evaluation set for the \textsc{en-el} language-pair.
\begin{table}[!t]
    \centering
        \scalebox{0.5}{
    \begin{tabular}{lll@{\hskip 0.2in}|@{\hskip 0.2in}lll}
        \toprule
        \multicolumn{3}{c}{\textbf{Revised}} & \multicolumn{3}{c}{\textbf{Original}}  \\[1pt]
        \toprule
        αστεροειδές & star	              & 	?       & απόστολος & apostolos    & \xmark \\ 
        προσφέρεται & offers	          &  \cmark	    & βραχνό & raucous         &  \xmark \\  
        κεραυνός & keravnos	              &  \xmark     & μπανζούλ & bangaon       & \xmark \\ 
        συμπυκνώνει & encapsulates	      & 	?       & βοηθητικές & auxiliary   & \cmark\\
        σεξτέτο & sexteto	              & \cmark 	    & ομιλήτρια & spokesperson & \cmark \\
        επιχειρηματολογία & argumentation & \cmark 	    & πρωτεργάτη & forerunner  &  \xmark \\
        επίπλωση & furniture	          & \cmark  	& αντιτρομοκρατική & anti-terrorist & \cmark \\
        μπουγκ & bug	                  & \xmark 	    & πλεκτά            & sweaters      & \cmark \\ 
        σχετικοί & related	              & \cmark 	    & εμβολιαστεί & vaccinated &    \cmark \\
        δορυφόρους & moons	              & \xmark 	    & αταξινόμητες & unclassified &  \cmark \\
        δειλή & timid	                  & \cmark	    & στέιν & steen                 & \xmark \\ 
        χάντινγκτον & huntingdon	      & \cmark 	    & χιλιοστό & millimeter & \cmark \\
        ποσότητες & amounts	              & \cmark	    & σελεστίν & célestine & \cmark \\
        πλακέ & squamous	              & \cmark 	    & κόβατς & kovács & \xmark\\
        αποποίηση & relinquishing	      & ?       	& σεμίνα & omni & \xmark \\
        ατμούς & vapors	                  & \cmark 	    & σπάιντερμαν & spider-man & \cmark \\
        τερματισμοί & endings	          & \cmark  	& πάνω & over & \cmark \\
        αλεξανδρινό & alexandrine	      & \cmark  	& ενδιαφέρων & love & \xmark \\
        σπασμοί & fits	                  & ? 	        & αγριόγατες & cats &      \xmark \\ 
        σίδερα & sidelines	              & \xmark	    & αγορα & trade & \cmark \\
        συνοδεύονται & are	              & \xmark 	    & επικεφαλίδα & header & \cmark \\ 
        διανέμονται & are	              & \xmark  	& μάσλοου & khan & \xmark \\
        θραύση & fracturing	              & \cmark  	& τεχνητά & artificially & \cmark \\
        κυβερνά & rule	                  & \cmark  	& πέτροβιτς & petrović & \cmark \\
        συνάξεις & meetings	              & \cmark  	& ανθίζει & flowers & \cmark \\
        χριστιανία & christianity	      & \cmark  	& ζήτω & vive & \xmark \\
        απειλούνται  & are	              & \xmark 	    & τυλίγει & picks & \xmark \\
        ποινικοποίηση & penalize	      & \cmark 	    & μπαέζ & ross & \xmark \\
        στερέωμα & stardom	              & \xmark 	    & φιλοδοξεί & is & \xmark \\
        τζαπ & elford	                  & \xmark 	    & τρυφερή & loving & ? \\
        ταυρομαχία & bullfighting	      & \cmark 	    & σωρός & remains & \xmark \\
        χειρός & handbags	              & ? 	        & χαλυβουργεία & works & \xmark \\
        κδ & cd	                          & ? 	        & μάιρα & chloe & \xmark \\
        τρομοκρατεί & terrorizes	      & \cmark 	    & συγκλόνισε & shocked & \cmark \\ 
        μακέι & mackey	                  & \cmark 	    & άτακτη & mischievous & \cmark \\
        ζάκυνθο & zakynthos	              & \cmark 	    & οταν & after & ?  \\
        συμπτωματολογία & symptomology	  & \cmark	    & εντομοφάγα & insectivores & \cmark \\
        πολυφυλετική & polyphyletic	      & \cmark 	    & κραδασμούς & vibrations & \cmark \\
        κούνια & cunha	                  & \xmark 	    & μπελάς & nuisance & \cmark \\
        καταβεβλημένος & overcome	      & \cmark 	    & πάστες & pastries & \cmark \\
        απάτες & scams	                  & \cmark 	    & διασπαστική & divisive & \cmark \\
        γιάννη & giannis	              & \cmark 	    & κατάληψη & capture & \xmark \\
        δηλητηριάσεις & poisonings	 & \cmark	   & παραδίδονται & surrender & \cmark \\
        φιλόξενοι & colorful	     & \xmark 	   & κλήρων & clergy & \cmark \\
        φημισμένος & renowned	     & \cmark      & σκεύη & vessels & \cmark \\
        φουσκωμένα & filled	         & ? 	       & λεπτονίων & leptons & \cmark \\
        υπονοούμενα & undertones	 & \cmark	   & εξάγονται & are & \xmark \\
        όριο & boundary	             & \cmark 	   & απότομο & abrupt & \cmark \\
        χαλάρωσε & relaxed	         & \cmark 	   & παρασυμπαθητικό & sympathetic & ? \\
        αισθητικός & aesthetic	     & \cmark 	   & ταρίχευση & embalming & \cmark \\
        ταμαντούα & tamanduas	     & \cmark 	   & κεκτημένο & precedent & \xmark \\
        εστίες & foci	             & ?	       & καλκούτα & kolkata & \cmark \\
        θεωρείται & is	             & \xmark 	   & σίρι & sirri & \cmark \\
        κορμό & trunk	             & \cmark 	   & ξεπερασμένο & obsolete & \cmark \\
        σπύρο & spyros	             & \cmark 	   & ανώμαλος & bumpy & \cmark \\
        αναισθητικά & anesthetics	 & \cmark	   & εξισορρόπησης & substance & \xmark \\
        στρατηγικές & strategic	     & \cmark 	   & πολυσακχαρίτης & polysaccharides & \cmark \\
        αναπνέει & breathe	         & \cmark 	   & επίπονες & persistent & \cmark \\
        εξουδετερώσει & neutralize	 & \cmark 	   & αμφιθέατρο & amphitheatre & \cmark \\
        μελαγχολική & melancholic	 & 	\cmark     & αναπληρωματικό & an & \xmark \\
        θυμήθηκε & recalled	         & \cmark 	   & εντελώς & entirely & \cmark\\
        πασχαλίτσα & ladybird	     & \cmark 	   & λιθόστρωτο & cobbled & \cmark \\
        πυροκροτητές & caps	         & ? 	       & διοικητικοί & administrative & \cmark \\
        κραυγαλέα & screaming	     & ? 	       & κομιστής & bearer & \cmark \\
        μολδαβίας & moldavia	     & \cmark 	   & συλλογικότητες & competitions & \xmark \\
        σαλιγκάρι & shilling	     & \xmark 	   & χουλιγκανισμού & micromanagement & \xmark \\
        ενισχυθεί & enhance	         & \cmark 	   & τσάρους & tsars & \cmark \\
        πρεσβυτέριο & presbytery	 & \cmark	   & ντόνελ & dorff & \xmark \\
        μάγιστρος & master	         & \cmark 	   & κίραν & kiran & \cmark \\
        αλτ & alt	                 & \cmark 	   & πρωτοποριακή & pioneering & \cmark \\
        χρονολογία & date	         & \cmark 	   & λένοξ & brookline & \xmark \\
        κανένα & any	             & \cmark 	   & λείπουν & are & \xmark \\
        κορμός & road	             & \xmark 	   & εξάντα & astronomy & \xmark \\
        καθαριστήριο & cleanup	     & \xmark 	   & πτωτική & downward & \cmark \\
        ανατεθεί & assigned	         & \cmark 	   & αρχιτεκτονικές & architectural & \cmark \\
        εξοικονόμηση & save	         & \cmark 	   & γαλλόφωνο & french-speaking & \cmark \\
        μπαρακούντα & barracudas	 & \cmark	   & μέντε & mede & \xmark \\
        ταυτοποίησης & identification	 & \cmark	   & εκθρονίζοντας & deposing & \cmark \\ 
        \toprule
    \end{tabular}}
    \caption{
    Manually labeled acceptability judgments for random $80$ error cases made by lexicons induced using the  original and revised bitexts. \cmark and {\xmark} 
    denote acceptable and unacceptable translation, respectively. ? denotes word pairs that may be acceptable in rare or specific contexts.
    }
    \label{tab:ground_truth_bli}
\end{table}

%
\section{Streamlining equivalization}
Based on ablation analysis presented in Table~\ref{tab:en2el_bleu} the best equivalization strategies consider candidates from two \nmt models trained independently to translate in opposite directions. In Table~\ref{tab:speeding_up} we show how our approach yields comparable results by replacing the two uni-directional models (\textsc{uni-nmt}) with a single bi-directional model (\textsc{bi-nmt})
while reducing training by $\sim30\%$.

\label{sec:appendix}

\end{document}